\documentclass{article}

\usepackage{microtype}
\usepackage{graphicx}
\usepackage{booktabs} 
\usepackage{subcaption}

\usepackage{hyperref}

\usepackage[accepted]{icml2023}

\usepackage{amsmath}
\usepackage{amssymb}
\usepackage{mathtools}
\usepackage{amsthm}

\usepackage[capitalize,noabbrev]{cleveref}

\theoremstyle{plain}

\theoremstyle{definition}

\theoremstyle{remark}

\usepackage[textsize=tiny]{todonotes}

\icmltitlerunning{Neural Regression For Scale-Varying Targets}

\begin{document}

\twocolumn[
\icmltitle{Neural Regression For Scale-Varying Targets}




\begin{icmlauthorlist}
\icmlauthor{Adam Khakhar}{a}
\icmlauthor{Jacob Buckman}{b}
\end{icmlauthorlist}

\icmlaffiliation{a}{University of Pennsylvania, USA}
\icmlaffiliation{b}{Montreal Institute for Learning Algorithms, Canada}

\icmlcorrespondingauthor{Adam Khakhar}{ak@alumni.upenn.edu}

\icmlkeywords{Machine Learning, ICML}

\vskip 0.3in
]



\printAffiliationsAndNotice{}  

\begin{abstract}
In this work, we demonstrate that a major limitation of regression using a mean-squared error loss is its sensitivity to the scale of its targets. This makes learning settings consisting of target's whose values take on varying scales challenging. A recently-proposed alternative loss function, known as histogram loss \citep{imani2018improving}, avoids this issue. However, its computational cost grows linearly with the number of buckets in the histogram, which renders prediction with real-valued targets intractable. To address this issue, we propose a novel approach to training deep learning models on real-valued regression targets, \textit{autoregressive regression}, which learns a high-fidelity distribution by utilizing an autoregressive target decomposition. We demonstrate that this training objective allows us to solve regression tasks involving targets with different scales.
\end{abstract}

\section{Introduction}
Regression is a central problem setting in the field of machine learning. In this setting, the goal is to learn a function mapping from some input space to a real-valued target from a dataset \citep{bishop2006pattern}. The modern approach to solving this family of tasks is deep learning \citep{goodfellow2016deep}, which canonically involves minimization of some loss via gradient descent. For regression tasks, the mean squared error (MSE) between the network's outputs and the target is a common choice due to its simplicity and convexity \citep{lehmann2006theory}. More recently, mean-absolute error (MAE) has also been introduced as a loss function for deep neural networks \citep{mae}.

Regression tasks with targets at varying scales constitute an important and challenging subset of problems. For example, consider the task of predicting the half-life of isotopes from their chemical structure. The targets for this task occupy more than 50 orders of magnitude, from Hydrogen-5 at $10^{-24}$ seconds to Tellurium-28 at $10^{30}$ seconds \citep{kondev2021nubase2020, barelyradio}. Intuitively, it is clear that the relative accuracy of predictions made on the rapidly-decaying elements would be almost irrelevant when measuring MSE as compared to the elements with large half-life. An algorithm trained to minimize MSE could simply predict 0 on all of these elements with smaller half-lives. Many realistic regression tasks exhibit targets with varying scales of this sort, albeit often to a less dramatic extent.

A common approach to addressing mis-scaled targets is output normalization \citep{shanker1996effect}. This solution transforms all targets in the training set to have a mean of $0$ and standard deviation of $1$. This technique is useful when targets are of a consistent scale, but that scale is either too large or too small. However, it does not address the issue of various targets of dramatically different scales.

Many recent works have investigated alternative losses with desirable properties \citep{huber2011robust,ghosh2017robust, imani2018improving,barron2019general,qi2020mean}, some of which alleviate this issue. In particular, \citet{imani2018improving} propose the histogram loss (HL). This approach divides the output range into discrete, non-overlapping ``bins'', converts each real-valued target into a distribution over bins, and minimizes the KL divergence \citet{kullback1951information} between the model's prediction and the target distribution. \citet{imani2018improving} argue that this choice of loss makes optimization easier due to its well-behaved gradients, and demonstrate empirically that this leads to improved generalization for certain choices of target distribution on a variety of tabular regression tasks. Follow-up study by \citet{gonzalez2022distributional} finds that this algorithm represents an improvement in all conditions on a vision task. Additional evidence justifying the usefulness of this loss can be found in deep reinforcement learning, where minimizing a histogram loss \citep{bellemare2017distributional} instead of MSE \citep{dqn} is found to improve performance.

However, the approach of \citet{imani2018improving} has a crucial limitation. Discretization of the target space introduces a trade-off between range, precision, and computation, since memory usage grows linearly with the number of buckets in the histogram. Additionally, the discretization of the target space makes predicting real-valued targets with precision infeasible. When targets vary across many orders of magnitude, as in our motivating example, memory limitations prevent us from capturing all of these values at the requisite fidelity. To resolve this issue, we propose a training objective called \textit{autoregressive regression}. In this technique, each bucket is represented as a sequence of class-tokens, which is predicted autoregressively. This allows us to represent an exponential (in the sequence length) number of buckets with a constant memory usage. The training objective, autoregressive regression, can be minimized using any neural architecture with an autoregressive head, including the Transformer architecture \citep{vaswani2017attention}.

\begin{figure*}[h]
    \begin{subfigure}{0.15\textwidth}
    \centering
        \includegraphics[width=\textwidth]{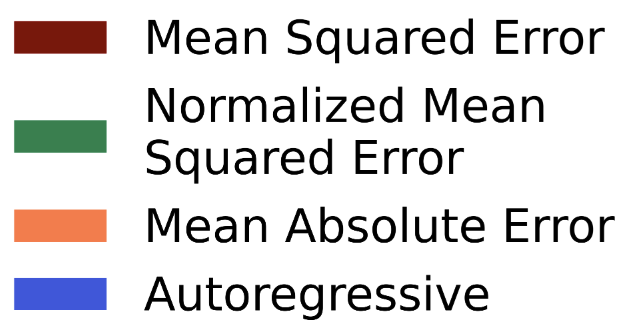}
    \end{subfigure}
    \hfill
    \begin{subfigure}{0.4\textwidth}
    \centering
        \includegraphics[width=\textwidth]{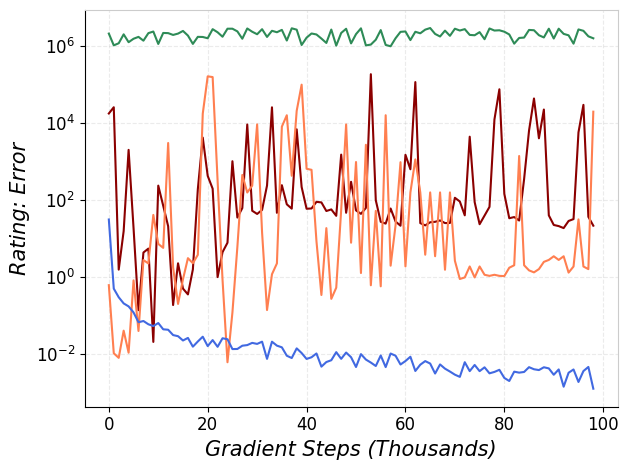}
        \caption{Test-set error of samples having a target denoting review Rating: smaller target scale.}
        \label{fig:review-rating}
    \end{subfigure}
    \hfill
    \begin{subfigure}{0.4\textwidth}
        \centering
        \includegraphics[width=\textwidth]{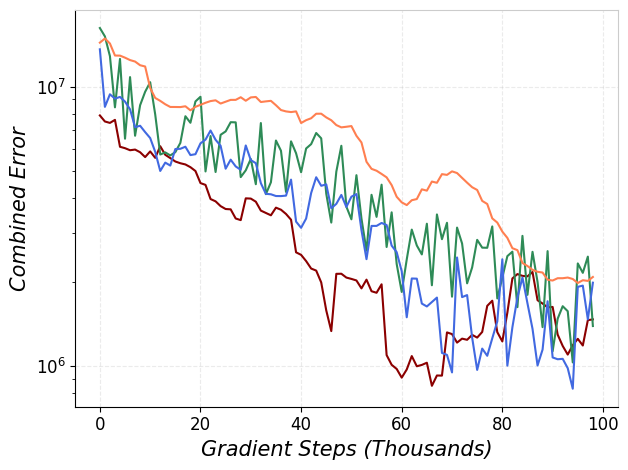}
        \caption{Total test-set error of all samples: including both Rating and Number of characters.}
        \label{fig:review-combined}
    \end{subfigure}
    \hfill
    \caption{Figure \ref{fig:review-combined} is the test-set error during training of the Amazon Review dataset task with a single target derived from two subtasks: review Rating and review Number of Characters. The input space (sequence of byte pair encodings) is concatenated with a binary indicator variable indicating if the target is the Rating of the review or the Number of Characters in the review. Figure \ref{fig:review-rating} is the test-set error of the subset of samples having a target that is the review Rating (smaller scale) during the same training process.}
    \label{fig:amazon-review}
\end{figure*}

In the Background section, we describe existing approaches, and in the Methods section, we explain our new technique. Finally, in the Experiments section, we present results from three sets of experiments selected to highlight our approach. First, we compare models trained using MSE, MSE with target normalization, and MAE to autoregressive regression models trained using the histogram loss on various tasks with multiple target scales, and see that only autoregressive regression consistently learns to predict targets at all scales. We further show that unlike MSE and MAE, our method has a stable optimal learning rate across all scales. Finally, we show that autoregressive regression can be tractably scaled up to support millions of buckets, whereas a na\"ive bucketing scheme (histogram loss) quickly encounters memory limits. 

\section{Background}
\label{sec:background}
In this section, we introduce the mean squared error (MSE), mean absolute error (MAE), and Histogram Loss (HL), the three major families of loss function that we consider in this work. We also introduce target normalization.

\subsection{Supervised Learning}
The problem setting we will be primarily studying is that of supervised learning. The goal is to choose the function $\hat{f}_\theta : \mathcal{X} \to \mathcal{Y}$, parameterized by $\theta \in \Theta$, that most closely resembles some ground-truth function $f : \mathcal{X} \to \mathcal{Y}$, based on a dataset $D$ of input-output pairs $D = \langle (x_0,y_0), (x_1,y_1), ..., (x_{|D|},y_{|D|}) \rangle$ where each $x_i \in \mathcal{X}$ and each $y_i = f(x_i)$. Of particular interest in this work are tasks where the targets $y_i$ exist across many scales. It is common to select the parameters which minimize some \textit{loss function}, $\mathcal{L} : (\mathcal{Y} \times \mathcal{Y}) \to \mathbb{R}$, averaged over the dataset:
\begin{equation}
    \min_{\theta \in \Theta} \frac{1}{|D|} \sum_{i=1}^{|D|}\mathcal{L}(f_{\theta}(x_i), y_i)
\end{equation}
When training neural networks, we approximate these optimal parameters using \textit{stochastic gradient descent}, in which we initialize the parameters at some $\theta_0$, and then iteratively modify them to follow a stochastic approximation of the gradient of the loss, as computed on a randomly-selected minibatch $B_t \subset D$ for each step of training $t$.
\begin{equation}
\begin{aligned}[c]
    J_t(\theta) := \frac{1}{|B_t|} \sum_{i=1}^{|B_t|}\mathcal{L}(f_{\theta}(x^{B_t}_i), y^{B_t}_i)
\end{aligned}
\qquad
\begin{aligned}[c]
    \theta_{t+1} := \theta_t - \alpha \frac{\partial}{\partial \theta_t}J(\theta_t)
    \label{graddesc}
\end{aligned}
\end{equation}
where $\alpha > 0$ is the learning rate.

\subsection{Mean Squared Error}
Mean squared error, also known as $\ell_2$ loss, is the canonical loss used for regression where the output space $\mathcal{Y} \subseteq \mathbb{R}$.
\begin{equation}
    \mathcal{L}_{\textsc{MSE}}(v, u) := (v - u)^2
\end{equation}
MSE has a probabilistic interpretation. We can consider our function $\hat{f}_{\theta}$ as predicting the mean of a Gaussian distribution $\mathcal{N}(\mu=\hat{f}_{\theta}(x), \sigma^2)$ for some arbitrary fixed variance $\sigma^2$. Minimizing the squared error is equivalent to choosing the distribution that maximizes the likelihood of the data.

\subsection{Mean Absolute Error}
Mean absolute error is also known as $\ell_1$ loss. It can be used for regression where the output space $\mathcal{Y} \subseteq \mathbb{R}$.
\begin{equation}
    \mathcal{L}_{\textsc{MAE}}(v, u) := |v - u|
\end{equation}
MAE can be interpreted as an error modeled by a Laplacian distribution \cite{mae}. Recent experiments have shown that deep neural network based regression optimized with the MAE loss function can achieve lower loss values than those obtained with MSE \cite{mae}.

\subsection{Histogram Loss}
In the task of learning some function $f : \mathcal{X} \to \mathbb{R}$, instead of directly predicting $y \in \mathbb{R}$, we can instead project $y$ to a particular point in the space of distributions over $\mathbb{R}$, which we denote $p(\mathbb{R})$. Histogram loss restricts this target distribution to be a histogram density, where the domain $\mathcal{Y} \subset \mathbb{R}$ is partitioned into $k$ bins of uniform width $w$, where $|\mathcal{Y}| = w*k$. Let projection function $g : \mathcal{Y} \rightarrow [0, 1]^k$ return the $k$-dimensional vector of probabilities that the target is in that bin. Thus, the projected target is a normalized histogram with density values $\frac{g(f(x))_i}{w}$ for each bin indexed by $1 \leq j \leq k$. To predict this target, our estimator must also output a distribution, $\hat{f}_\theta : \mathcal{X} \to p(\mathcal{Y})$. The histogram loss is defined as the cross entropy between the distributions $\hat{f}_\theta(x)$ and $g(f(x))$:
\begin{equation}
    \mathcal{L}_\textsc{HL}(p, q)=-\sum_{i=1}^k p_{j}\textrm{log}q_{j}
\end{equation}
Note that, just as we saw in the case of MSE, this can be interpreted as a maximum-likelihood estimate of the data.

\begin{figure*}
    \begin{subfigure}{0.15\textwidth}
    \centering
        \includegraphics[width=\textwidth]{figures_v2/color_legend.png}
    \end{subfigure}
    \hfill
    \begin{subfigure}{0.4\textwidth}
    \centering
        \includegraphics[width=\textwidth]{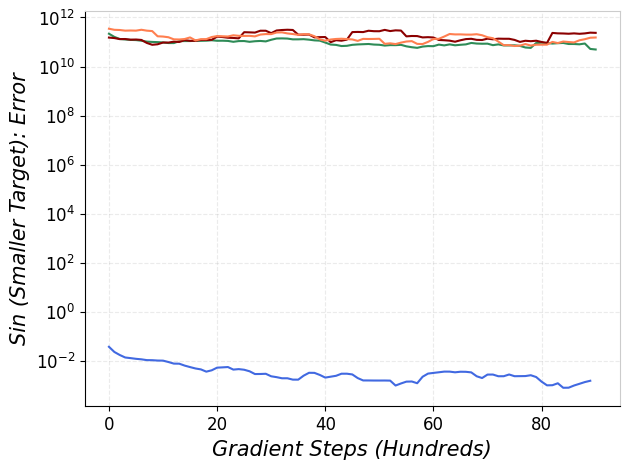}
        \caption{Error of samples with $\sin$ target: smaller target scale.}
        \label{fig:one-dim-sin-s}
    \end{subfigure}
    \hfill
    \begin{subfigure}{0.4\textwidth}
        \centering
        \includegraphics[width=\textwidth]{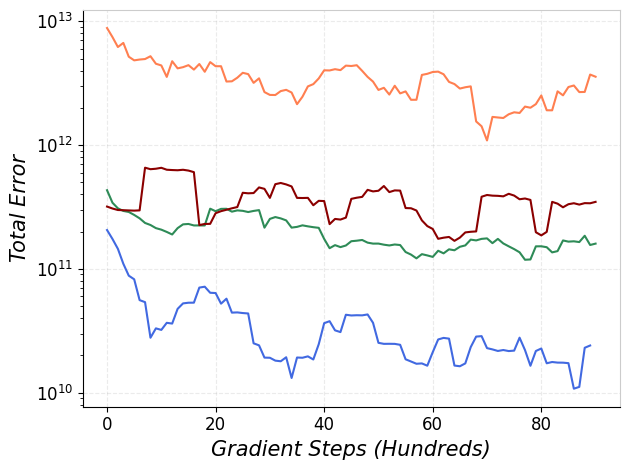}
        \caption{Total error of all samples including targets from both $\log$ and $\sin$ functions.}
        \label{fig:one-dim-sin-s-log-l}
    \end{subfigure}
    \hfill
    \caption{Figure \ref{fig:one-dim-sin-s-log-l} is the total test-set error of the 1-dimensional toy dataset regression task with a single target derived from two subtasks: $\log$ function with a mean value of $9 \times 10^6$ and $\sin$ function with a mean value of $1$. The one dimensional input is negative if the target is the $\sin$ function and positive if the target is the $\log$ function. Figure \ref{fig:one-dim-sin-s} is the test-set error of the subset of samples having a target that is the $\sin$ function (smaller scale) during the same training process.}
    \label{fig:algo-sin-s-log-l}
\end{figure*}

\citet{imani2018improving} describe several choices for the projection function $g$. These include a truncated and discretized Gaussian distribution (HL-Gaussian), a delta distribution resulting in a single bin of 1 (HL-OneBin), and an interpolation between a delta distribution and the uniform distribution (HL-Uniform). \citet{imani2018improving} provide experimental evidence that HL-Gaussian had the best generalization performance out of these techniques on a set of tabular benchmarks. However, in this work, we instead focus on HL-OneBin, due to its simplicity; since there is no ambiguity, we refer to this technique simply as HL. Concretely, we have
\begin{equation}
    g(y)_j = 
\Big\{
    \begin{array}{ll}
        1 & \text{if } \frac{j}{k} \leq \frac{y - \inf \mathcal{Y}}{\sup \mathcal{Y} - \inf \mathcal{Y}} < \frac{j+1}{k}, \\
        0 & \textrm{otherwise.}
    \end{array}
\end{equation}

\subsection{Target Normalization}
In the supervised learning setting, where the goal is to learn some ground-truth function $f : \mathcal{X} \to \mathcal{Y}$, for $y_i \in \mathbb{R}$, output normalization refers to transforming $\mathcal{Y}$ to $\mathcal{Y'}$ such that $E[\mathcal{Y'}] = 0$ and $Var[\mathcal{Y'}] = 1$. This is accomplished by computing the mean and standard deviation of the $\mathcal{Y}$ in the training set of data, and setting 
\begin{equation} \label{eqn:output_norm}
    y'_i := \frac{y_i-E[\mathcal{Y}]}{\sigma_{\mathcal{Y}}}
\end{equation}
After training, the inverse of equation \ref{eqn:output_norm} is used to convert the prediction $y_i'$ to $y_i$. Output normalization is usually performed when the target, $\mathcal{Y}$, takes values at very large or small scales.

\subsection{Limitations of Existing Approaches}
We now discuss limitations of these losses with respect to our motivating task: regression on datasets whose targets vary in scale. Each of these techniques has a crucial drawback and, as a result, none of the three approaches considered in prior literature is a feasible solution. Additionally, we describe out output normalization fails in cases where the target is a mixture of distributions, where the distributions have standard deviations at different scales. These limitations motivate the algorithm we propose in the Methods section, and are validated empirically in the Experiments section.

\subparagraph{Scale Sensitivity of MSE and MAE.}

When mean squared error is used as the loss function, the scale of the targets impacts the scale of the gradient. If a prediction $\hat{y}$ and its target $y$ are both scaled by a factor $z > 1$, the MSE will grow to $z^2(y - \hat{y})^2$ and, crucially, its derivative will grow to $2z^2(y - \hat{y})$. When two targets $y_1, y_2$ have dramatically different scales, $|y_1| \ll |y_2|$, the loss on the second target will be far larger, $(y_1 - \hat{y}_1)^2 \ll (y_2 - \hat{y}_2)^2$, even when the relative errors are similar, $\frac{|y_1 - \hat{y}_1|}{y_1} \approx \frac{|y_2 - \hat{y}_2|}{y_2}$. This gap is reflected in the scale of the gradient, leading to SGD primarily updating the parameters in a direction which improves the prediction on the larger-scaled target. Similarly, when mean absolute error is used as the loss function, only the larger targets will incur loss with a consistent sign. In situations where the model is limited (by scale or compute) in its ability to represent the target function, this can lead to poor-quality predictions on smaller-scaled targets. MSE and MAE are therefore limited in their usefulness on tasks with targets at multiple scales.

\subparagraph{Computational Tractability of Basic HL.}
The prototypical approach to classification using neural networks is to output one logit per class. This has an important drawback: as the number of classes increases, the memory usage increases linearly. It is easy to see that it is intractable to make predictions over a large number of classes using this approach. For example, if the goal is to learn some function $f : \mathcal{X} \to \mathcal{Y}$, where $|\mathcal{Y}| = 1000000$, and we use a neural network whose final hidden layer has dimension $4096$ with a 32-bit float representation, then the parameters of the final layer alone will be approximately 16 gigabytes, larger than the memory of many commercial GPUs.

This represents a challenge when using the histogram loss
with deep learning, in particular for regression tasks that consist of multiple subtasks of varying magnitude. Consider a regression task where targets range from $(-10000, 10000)$, and two significant figures of precision are desired; it is easy to see that this translates into a 1000000-class classification problem. The central contribution of this work is to propose a method which can tractably solve such problems.

\subparagraph{Failure of Target Normalization on Targets with Varying Scale}
Consider a target whose value takes on a function $f_a:  \mathcal{X}_a \to \mathcal{Y}_a$ on some space of inputs $\mathcal{X}_a$ and $f_b:  \mathcal{X}_b \to \mathcal{Y}_b$ on some space of inputs $\mathcal{X}_b$, such that $\mathcal{X}_a \cap \mathcal{X}_b = \emptyset$ and $\mathcal{Y}_a \cup \mathcal{Y}_b = \mathcal{Y}$. Next, consider the case where $\sigma_{\mathcal{Y}_a} >> \sigma_{\mathcal{Y}_b}$. The normalized target is proportional to the inverse of the standard deviation of the target times the target: $\mathcal{Y}' \propto \frac{1}{\sigma_\mathcal{Y}}\cdot \mathcal{Y}$. Because normalization is computed using the training set target mean and standard deviation for samples from all functions $f_a$ and $f_b$, the standard deviation of the target space is much greater than the standard deviation of the targets resulting from $f_b$: $\sigma_\mathcal{Y} >> \sigma_{\mathcal{Y}_b}$. Thus the normalized targets $y_{i, b}'$ for samples computed from $f_b$, will have a much smaller variance as compared to $y_{i, a}'$ for samples computed from $f_a$. The effect of this difference in variance is that the model is encouraged to learn fine-tuned differences in targets from $f_a$, having targets at larger scale, and not in $f_b$, which has targets at smaller scale.


\section{Method}
\label{sec:algo}
In this section we introduce our proposed technique, autoregressive regression. The key insight motivating this approach is that any number can be uniquely decomposed into a sequence of coefficients using an exponential basis. This idea is ancient and well-understood; it underpins the Arabic numerals ubiquitous in modern mathematics. Our contribution is to show how this basic idea can be applied to deep learning in order to construct a tractable algorithm for regression at multiple scales.

As described in the Background section, a model trained using the histogram loss must predict a distribution over $k$ bins. In prior work \citep{imani2018improving}, this was done by constructing a network to directly output a vector of $k$ probabilities. This can be interpreted as predicting a probability distribution over 1-digit numbers in base $k$. We propose to generalize this approach to other choices of base: the same object can be instead described as a distribution over $s$-digit numbers in base $b := \sqrt[s]{k}$, where any number $0 \leq a < k$ is represented as a list $\dot{a}$ whose components $\dot{a}_j := \lfloor \frac{a \mod b^{j+1}}{b^j} \rfloor$ for $0 \leq j < s$ correspond to the coefficients of each power, $a = \sum_{j=0}^k \dot{a}_i b^j$.

We can use the chain rule of probability to implicitly specify a distribution over these lists from their conditional probabilities, $p(a) = \Pi_{j=0}^{k} p(\dot{a}_j \mid \dot{a}_{<j})$. We can therefore represent the distribution of probabilities using an autoregressive neural network, using standard tools for neural sequence modeling. Our neural function approximator looks at both the input $x$ and a partial sequence of already-observed tokens, and output a vector of $b$ probabilities, representing the distribution of possible values of the next token. A prediction is made for each prefix of the sequence, and the resulting log-probabilities are summed to produce the overall log-probability required for the computation of cross-entropy. This is precisely equivalent to the well-studied task of conditional generative modeling, such as translation \citep{yang2020survey}; all techniques from that literature can be applied to our domain. In this work, we utilize the powerful and popular Transformer model \citep{vaswani2017attention}.

By construction, we have $k = b^s$, so the number of buckets we can represent grows exponentially with the number of autoregressive steps taken. This relationship is where our approach derives its power. For a fixed output space $\mathcal{Y}$, high-fidelity predictions require a large number of buckets $k$. With our approach, this is typically tractable in just a small number of steps, since exponential functions grow very quickly. In contrast, prior approaches \citep{imani2018improving} which only consider $s=1$ can achieve only linear scaling of $k$ (by increasing $b$), potentially exhausting memory before sufficient fidelity can be reached.


\section{Experiments}
\label{sec:exp}
We hypothesize that our proposed method, autoregressive regression, does not suffer from either of the limitations described above. To this end, we perform experiments which highlight the limitations of prior approaches, and show that autoregressive regression models trained to solve the same tasks do not suffer from these issues. Experiments are performed on two toy datasets, a 1-D prediction task and the MNIST dataset \cite{deng2012mnist}, as well as the real-world Amazon Review dataset \cite{amazonreview}.\footnote{Open-source PyTorch code to replicate all experiments is available at \url{https://github.com/anonymized}.}

\subsection{Learning At Multiple Scales}

These experiments compare the empirical performance of models trained using MSE, MSE with target normalization, and MAE to autoregressive regression models using the histogram loss. In particular, we investigate the ability to learn subtasks with targets at different scales and sensitivity to learning rate.

Note that although our experiments use autoregressive regression, the improvements in this section do not stem from this choice directly, but rather from the use of histogram loss. In that regard, these experiments build on the work of \citet{imani2018improving} in motivating the use of histogram loss for regression, and are complimentary. While that work demonstrated that certain variants of the histogram loss could lead to improved generalization, the goal of our experiments is to investigate its performance in the specific setting of regression of targets at multiple scales.

\begin{figure*}[h]
    \begin{subfigure}{0.15\textwidth}
    \centering
        \includegraphics[width=\textwidth]{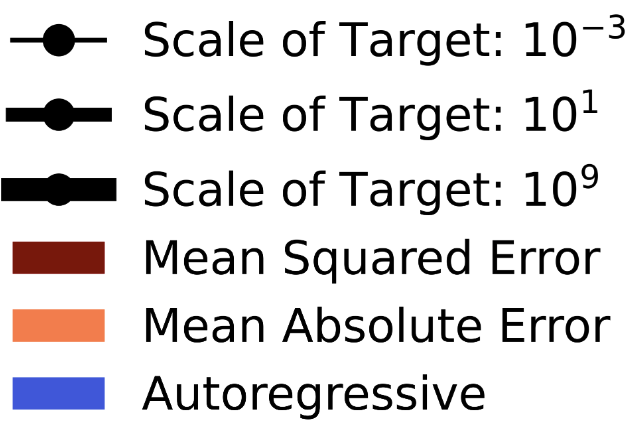}
    \end{subfigure}
    \hfill
    \begin{subfigure}{0.27\textwidth}
    \centering
        \includegraphics[width=\textwidth]{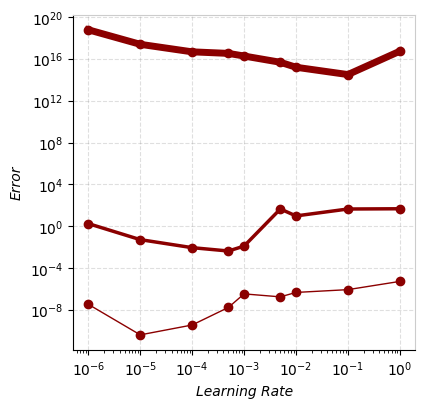}
        \caption{Models trained with mean squared error.}
        \label{fig:lr-mse}
    \end{subfigure}
    \hfill
    \begin{subfigure}{0.27\textwidth}
    \centering
        \includegraphics[width=\textwidth]{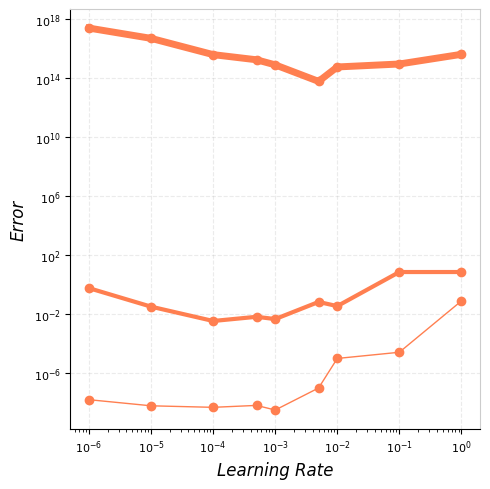}
        \caption{Models trained with mean absolute error.}
        \label{fig:lr-mae}
    \end{subfigure}
    \hfill
    \begin{subfigure}{0.27\textwidth}
        \centering
        \includegraphics[width=\textwidth]{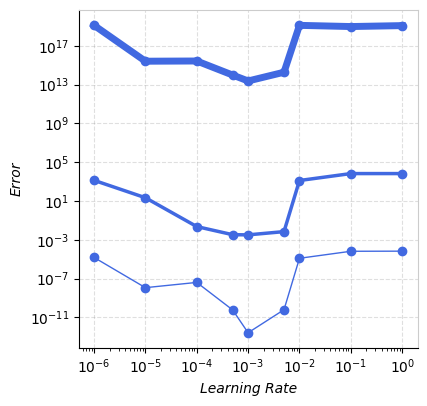}
        \caption{Models trained with autoregressive regression, using 10 tokens and 8 autoregressive steps.}
        \label{fig:lr-arr}
    \end{subfigure}
    \hfill
    \caption{Test-set error at convergence, after training MSE, MAE and autoregressive models on our MNIST domain at various learning rates with targets at various scales. Note that the optimal learning rate is stable only for autoregressive regression.}
    \label{fig:lr-combined}
\end{figure*}

\subsubsection{Multi-task Regression}
In these experiments, we empirically evaluate the hypothesis that MSE, MSE with target normalization, and MAE fail to learn small-scale targets when multiple target scales are present, whereas models using autoregressive regression (with histogram loss) are able to learn all targets regardless of scale. We conducted experiments in three domains of increasing complexity: a toy 1D task, a vision task based on MNIST \citep{deng2012mnist}, and a natural language task derived from Amazon review data \citep{amazonreview}.

For each domain, we construct a single target compromising of two subtasks of similar difficulty, but where the scale of one subtask is significantly larger than the other. Specifically, for each input, the target is either from subtask $\mathcal{A}$ or subtask $\mathcal{B}$, with equal likelihood. The input to the model is treated with a certain function $f_\mathcal{A}$ or $f_\mathcal{B}$ depending on if the target is derived from subtask $\mathcal{A}$ or subtask $\mathcal{B}$. For example, in the one-dimensional input case, samples with target from subtask $\mathcal{A}$ are all negative ($\mathcal{X}_{\mathcal{A}} \in \mathbb{R}^{-}$) while the samples with target from subtask $\mathcal{B}$ are all positive ($\mathcal{X}_{\mathcal{B}} \in \mathbb{R}^{+}$).
In our figures, we separate model evaluation on samples from each subtask and plot the error for the smaller subtask independently as well as the the combined error of both subtasks, in order to identify whether the model has learned to make accurate predictions on both.

\subparagraph{Toy 1D Domain.}
In this domain, the input consists of real numbers uniformly sampled from $[0, 1]$, and two simple target functions were used: $\sin$ and $\log$, but with the $\sin$ function scaled up by 6 orders of magnitude.

To avoid the potential confounder of generalization, we generate fresh data for each minibatch, evenly balanced between the two subtasks. Our architecture for this domain is a simple feedforward neural network \citep{widrownn}. Refer to appendix for additional experimental details.

Figure \ref{fig:algo-log-s-sin-l} depicts the results of this experiment, which show a clear pattern in support of our hypothesis: autoregressive regression solves both subtasks, whereas MSE, MSE with target normalization, and MAE solves only the subtask with targets at larger scale.
We validate that this result is due to the limitations we have identified by confirming that when two tasks are at the same scale, all techniques succeed to optimize all targets (Figure \ref{fig:algo-sin-s-log-s}); furthermore, when we switch which task is scaled up, MSE, MSE with target normalization, and MAE still fail to solve the task with smaller targets (Figure \ref{fig:algo-sin-s-log-l}).

\subparagraph{MNIST Domain.}
In this domain, each input is a 56x56 greyscale image constructed from the MNIST dataset \citep{deng2012mnist} by concatenating four digits uniformly sampled from the train set (see Figure \ref{fig:mnist-sample} in the Appendix for an example). The MNIST dataset \citep{deng2012mnist} is traditionally an image classification task; to use this domain for regression, we simply map each image to the decimal constructed by reading off each of its four digits (giving values approximately uniformly distributed between 0 and 1). Furthermore, in order to generate two subtasks, we again transform these values using the $\sin$ or $\log$, and scale up one of the two sets of targets by 6 orders of magnitude. The dataset for this experiment consists of all four-digit numbers constructed from the MNIST dataset. Our architecture for this domain is a convolutional neural network \citep{lecun1995convolutional}. See the Appendix for additional experimental details.

We see in Figures \ref{fig:mnist-sin-s-log-l} and \ref{fig:mnist-sin-l-log-s} that only the larger target is learned by MSE and MAE, whereas both subtasks are learned by autoregressive regression. When both targets are at the same scale, they are learned by all methods (Figure \ref{fig:mnist-sin-s-log-s}). 

\subparagraph{Amazon Review Dataset.}
Our final domain is a real-world natural language data set: Amazon Review Data \citep{amazonreview}. Given the first $500$ tokens of an Amazon Review, we choose two prediction targets for regression: (a) the star rating the reviewer assigned and (b) the total number of characters in the entire review. Note that because byte pair encoding \citep{bpe} was used to train the model and only the first $500$ tokens from the review were given as context to the model, predicting the total number of characters in the review is a non-trivial task. Additionally, note that the scale of the first target (the star rating) is significantly smaller than that of the second target (the review length). Our architecture for this domain is a Tranformer \citep{vaswani2017attention}. See the Appendix for additional experimental details.

We see in Figure \ref{fig:amazon-review} that only the subtask with larger targets, character count, is learned by MSE, MSE with target normalization, and MAE, whereas both are learned by autoregressive regression. In fact, the error for the rating subtask when trained by MSE appears to slightly \textit{increase} as more gradient steps are taken.

\subsection{Learning Rate Stability}
In the following experiments, we compare the effects of target scale on optimal learning rates between autoregressive regression, MSE, and MAE. We expect that since the gradient of the MSE and MAE loss are sensitive to the scale of its targets, optimal learning rates will be dependent on that scale. This means that learning rates need to be tuned on a per-task basis when training with MSE and MAE. In contrast, we hypothesize that because loss is scale invariant for autoregressive regression, optimal learning rates are stable across targets of various scales. To test this hypothesis, we experimented with various learning rates using the MNIST domain described above (see Figure \ref{fig:mnist-sample}), and just a single target corresponding to the $\sin$ transformation.

The results in Figure \ref{fig:lr-combined} confirm our hypothesis that, for models trained with autoregressive regression, optimal learning rate is stable across a wide range of scales, while the optimal learning rates for regression models trained with MSE vary significantly. For example, the optimal learning rate for the MSE model trained with targets at the scale of $10^9$ appears to be near $10^{-1}$, while optimal learning rate for the model trained with targets at the scale of $10^{-3}$ appears to be near $10^{-5}$. This result contrasts with the optimal learning rates for the model trained with autoregressive regression where the optimal learning rate appears to be around $10^{-3}$ at all scales.

\subsection{Computational Tractability}
\begin{figure}[t]
    \centering
    \includegraphics[width=3.5in]{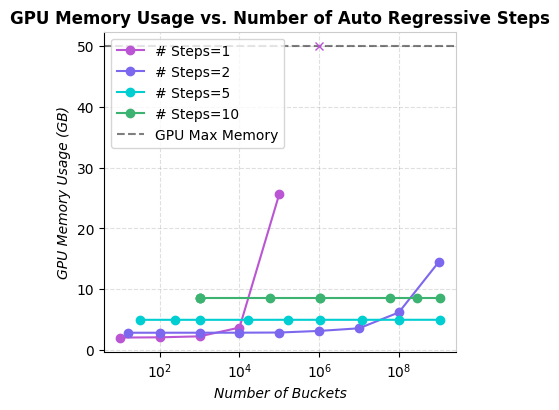}
    \caption{Memory usage over various number of autoregressive steps and various desired numbers of output classes. Step size of 1 is equivalent to histogram loss. The $\times$ indicates an out-of-memory error. Experiments were performed on RTX A6000 GPU.}
    \label{fig:memcombined}
\end{figure}

In these experiments, we aim to compare our proposed technique, autoregressive regression, with a more standard non-autoregressive approach to minimizing the histogram loss, e.g. from \citet{imani2018improving}. In particular, we claim that only autoregressive regression can tractably make predictions of arbitrarily high scale and precision. To test this hypothesis, we constructed a set of simple experiments using our MNIST domain, with various levels of desired fidelity. We trained models to minimize the histogram loss on this task and measured the memory usage using various numbers of autoregressive steps. Note that when using just a single autoregressive step, this is equivalent to the baseline histogram loss method proposed in \citet{imani2018improving}. The results in Figure \ref{fig:memcombined} confirm our hypothesis that using an autoregressive decomposition allows us to tractably scale to an arbitrarily large number of buckets, enabling use of the histogram loss in scenarios where predictions are required to be both high-fidelity and at a large scale.

\section{Discussion}

Our results clearly show the power of autoregressive regression in solving tasks with targets at different scales. In this work, we constructed experiments which emphasize this property, each consisting of a single target comprised of two subtasks whose targets are at dramatically different scales. However, note that this choice was merely illustrative. We expect very few real-world tasks to be so cleanly divisible. The value of autoregressive regression lies in the fact that many real-world tasks have targets of varying scales, despite the absence of clear delineations between subtasks. This is also the justification for why output standardization \citep{shanker1996effect}, an approach in which each subtask is standardized to have targets with mean zero and unit variance, does not solve our problem. Although normalizing each subtask's targets separately would indeed return them to the same scale and permit learning (see e.g. Figure \ref{fig:algo-sin-s-log-s}), we are primarily interested in approaches for solving problems where many subtasks are mixed together, and so per-subtask normalization is not possible.

For some real-world tasks, the scale sensitivity of MSE and MAE may be acceptable. For example, when predicting power usage of a datacenter \citep{prevost2011prediction}, it may be the case that the largest costs are accrued on days with enormous energy draw, and these are therefore the most important predictions. But there are also many tasks where the opposite is true. For example, mean arterial pressure \citep{demers2021physiology} is an important medical diagnostic metric. In healthy patients, this will be measured at 80-100; for at-risk patients, it can drop to as low as 60. These situations are dangerous, and small movements in mean arterial pressure become crucial to detect because they can translate to dramatically different clinical outcomes. A model using MSE or MAE to predict mean arterial pressure may do a poor job making predictions for patients whose pressure is in the 60s due to its smaller scale. Yet for this task, that is exactly where accurate predictions are most essential.

One limitation of autoregressive regression is that computing the predicted mean is computationally expensive. MSE and MAE models output the mean directly, and non-autoregressive HL models output the full vector of probabilities, and so the exact predicted mean can be computed cheaply via marginalization. However, in autoregressive regression, marginalizing exactly requires generating every possible output sequence, which is typically intractable. Instead, an approximate mean can be computed via sampling. Like other neural sequence models, drawing a batch of samples in parallel is relatively cheap, and this cost is only incurred at inference time in standard supervised learning tasks. However, this does pose a more serious challenge for its use in deep reinforcement learning, where computing the mean of each action is necessary at train time, due to the argmax in the Bellman optimality equation \citep{sutton2018reinforcement}.

\section{Related Work}

Our contributions build upon the work of \citet{imani2018improving}, who first introduced the histogram loss. \citep{imani2018improving} showed that a specific variant of the histogram loss, HL-Gaussian, improves generalization performance on a small set of tabular tasks. We show that even the simplest variant, HL-OneBin, is helpful in correcting some limitations of MSE, and we demonstrate this empirically on a variety of domains including tabular, images, and text. Also, our proposed algorithm removes a limitation of \citet{imani2018improving} by making HL tractable at any fidelity.

Concurrent work by \citet{li2022adacat} proposes a different solution to the challenge of high-fidelity histogram loss. The authors suggest using an adaptive categorization, where bucket widths are adjusted dynamically so that fidelity can be increased in parts of the distribution where it is relevant. This allows a fixed number of buckets to be used more efficiently. However, weaknesses of this approach include additional optimization complexity due to the non-differentiability of the bucket edges; it also does not address the case where high fidelity is required everywhere. Overall, this work is complimentary to our approach, and an interesting direction for future work is to explore how to use these two techniques in conjunction.

Another related line of work comes from generative modeling of images. In this field, images are often represented as an array of real numbers \citep{kingma2013auto}. However, some work such as PixelCNN \citep{van2016pixel} have shown that it is possible to construct powerful generative models by instead representing images as a sequence of discrete tokens. This transformation is similar to the one we propose, in that it involves replacing prediction of a real-valued target with that of a categorical distribution.

In deep reinforcement learning \citep{sutton2018reinforcement}, a neural network is often used to approximate a value function. Although this is not a supervised regression task due to bootstrapping and a constantly-changing dataset, it bears many similarities, in that an (often image) input is used to predict a real-valued target. Additionally, deep reinforcement learning is a problem setting where targets can have varying scales, and good predictions at all scales are important; this is a central challenge \citep{hessel2019multi}. One deep reinforcement algorithm known as C51 \citep{bellemare2017distributional} converts real-valued targets into discrete buckets and minimizes a cross-entropy, analogous to the histogram loss \citep{imani2018improving}. However, due to the issues with memory usage, that algorithm only uses a relatively low-fidelity prediction of 51 bins. Follow-up work by \citet{dann2021adapting} proposes an algorithm to increase the fidelity by decomposing the return into a sequence of exponential coefficients, similar to our approach; however, their algorithm predicts each coefficient independently rather than autoregressively, severely limiting the distributions that they can represent.

\section{Conclusion}
We introduced a novel training objective for regression, called autoregressive regression, which decomposes real valued targets into a sequence of distributions to predict one at a time using previously predicted distributions as context. We highlight that autoregressive regression enjoys the theoretical optimization properties of histogram loss while being capable of modeling arbitrarily-large targets to arbitrary precision. We demonstrated the effectiveness of autoregressive regression on a diverse set of domains, including tabular data, image data, and natural language data. In future work, we hope to further explore the practical relevance of autoregressive regression by evaluating this technique on more real-world domains of practical interest, and extending it to natural application areas such as distributional reinforcement learning.

\section*{Software and Data}
An open-sourced implementation is available at \url{https://github.com/adamkhakhar/autoregreg}.

\section*{Acknowledgements}
We would like to thank Professor Osbert Bastani and The Trustworthy Machine Learning Group at the University of Pennsylvania for their helpful discussions. We would also like to thank Professor Xi Chen from New York University for his comments.


\bibliography{refs}
\bibliographystyle{icml2023}

\newpage
\appendix
\onecolumn
\section{Appendix}
\subsection{Experiment Details}
\subsubsection{Toy 1D Domain}
\begin{figure*}[h]
    \begin{subfigure}{0.15\textwidth}
        \centering
            \includegraphics[width=\textwidth]{figures_v2/color_legend.png}
    \end{subfigure}
    \hfill
    \begin{subfigure}{0.4\textwidth}
        \centering
            \includegraphics[width=\textwidth]{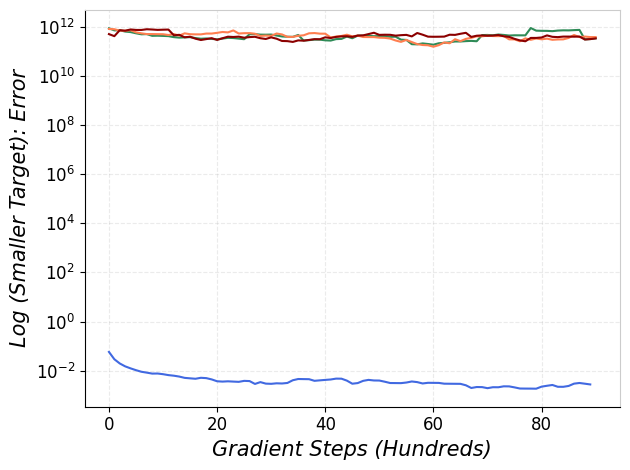}
        \caption{Error of samples with $\log$ target: smaller target scale.}
        \label{fig:one-dim-small-log-s}
    \end{subfigure}
    \hfill
    \begin{subfigure}{0.4\textwidth}
        \centering
        \includegraphics[width=\textwidth]{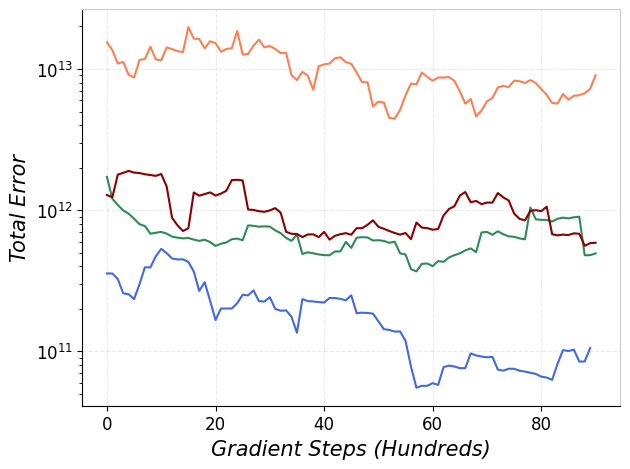}
        \caption{Total error of all samples including targets from both $\log$ and $\sin$ functions.}
        \label{fig:one-dim-log-s-sin-l-b}
    \end{subfigure}
    \hfill
    \caption{Figure \ref{fig:one-dim-log-s-sin-l-b} is the total test-set error of the 1-dimensional toy dataset regression task with a single target derived from two subtasks: $\log$ function with a mean value of $1.67$ and $\sin$ function with a mean value of $8 \times 10^6$. The one dimensional input is negative if the target is the $\sin$ function and positive if the target is the $\log$ function. Figure \ref{fig:one-dim-small-log-s} is the test-set error of the subset of samples having a target that is the $\log$ function (smaller scale) during the same training process.}
    \label{fig:algo-log-s-sin-l}
\end{figure*}

\begin{figure*}[h]
    \begin{subfigure}{0.15\textwidth}
    \centering
        \includegraphics[width=\textwidth]{figures_v2/color_legend.png}
    \end{subfigure}
    \hfill
    \begin{subfigure}{0.4\textwidth}
    \centering
        \includegraphics[width=\textwidth]{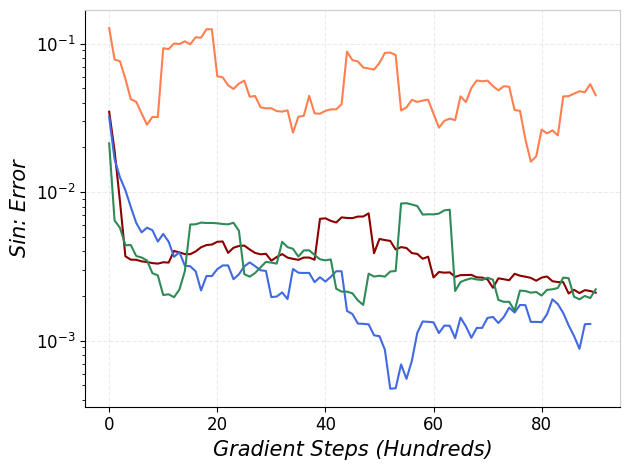}
        \caption{Error of samples with $\sin$ target.}
        \label{fig:one-dim-both-sin-s}
    \end{subfigure}
    \hfill
    \begin{subfigure}{0.4\textwidth}
        \centering
        \includegraphics[width=\textwidth]{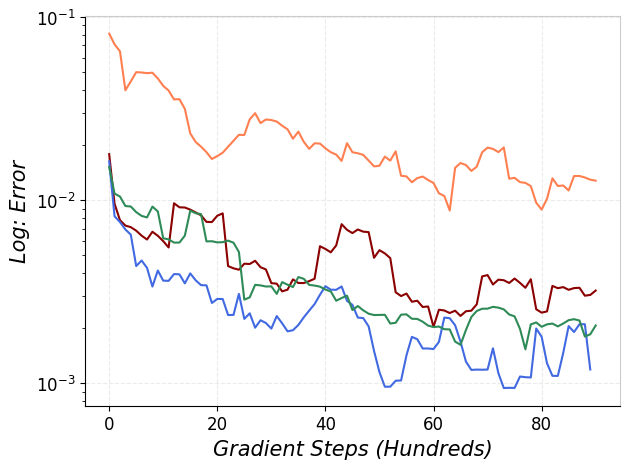}
        \caption{Error of samples with $\log$ target.}
        \label{fig:one-dim-both-log-s}
    \end{subfigure}
    \hfill
    \caption{The figure above depicts the test-set error during training of the 1-dimensional toy dataset regression task with a single target derived from two subtasks: $\log$ function with a mean value of $1.67$ and $\sin$ function with a mean value of $1$. The one dimensional input is negative if the target is the $\sin$ function and positive if the target is the $\log$ function. Figure \ref{fig:one-dim-both-sin-s} is the test-set error of the subset of samples having a target that is the $\sin$ function. Figure \ref{fig:one-dim-both-log-s} is the test-set error of the subset of samples having a target that is the $\log$ function during the same training process.}
    \label{fig:algo-sin-s-log-s}
\end{figure*}

In the following section, we describe the data, model, and optimization techniques for the experiments with the algorithmically generated 1-dimensional domain.
\subparagraph{Dataset}
The input to the model is a uniformly sampled float in the range (0, 1). All models were trained with $10,000,000$ samples with a batch size of $1,000$, resulting in $10,000$ gradient steps. We define the following 4 functions:
\begin{align}
    \sin_{s}(x) &= \sin(2\pi \cdot x) + 1 \\
    \sin_{l}(x) &= 3\cdot 10^6\cdot (\sin(2\pi \cdot x) + 1) + 5\cdot 10^6 \\
    \log_{s}(x) &= 2\cdot(\log(x+.4)+1) \\
    \log_{l}(x) &= 5\cdot 10^6\cdot(\log(x+.4)+1)+5\cdot 10^6
\end{align}
The experiment in Figure \ref{fig:algo-log-s-sin-l} corresponds to a model trained on subtasks $\log_s$ and $\sin_l$, the experiment in Figure \ref{fig:algo-sin-s-log-l} corresponds to a model trained on $\sin_s$ and $\log_l$, and the experiment in Figure \ref{fig:algo-sin-s-log-s} corresponds to a model trained on $\sin_s$ and $\log_s$. Samples with targets having a $\sin$ function were multiplied by $-1$ to have an input space be uniformly distributed from -1 to 0. Samples with targets having a $\log$ function were kept in its original form, being uniformly distributed from 0 to 1. In a single training batch, each sample has an equal probability of having a target be a $\sin$ or $\log$ function.

\subparagraph{Model and Optimization}
Each Figure (\ref{fig:algo-log-s-sin-l}, \ref{fig:algo-sin-s-log-l}, \ref{fig:algo-sin-s-log-s}) presents the error for the 2 subtasks over the course of training for a model trained with MSE, a model trained with MSE including target standardization, a model trained with MAE, and a model trained with autoregressive regression. All models were optimized via the Adam optimizer \citet{https://doi.org/10.48550/arxiv.1412.6980}. Learning rates for presented results were chosen using the same methodology for all types of models: all half orders of magnitude from $10^1 \rightarrow 10^{-6}$ were tested and the best performing learning rate was chosen. The autoregressive models were able to be trained with the same learning rate regardless of target scale, with a learning rate of $.005$, whereas the MSE and MAE models required varying learning rates dependent on the scale of the targets. The model architecture is a 5-layer feed forward network of dimension $1,024$ with rectified linear unit activation function \citet{https://doi.org/10.48550/arxiv.1803.08375}. The model trained with autoregressive regression has $8$ autoregressive steps each with a bin size of $100$. Training was executed on a RTX A6000 GPU. Each experiment was run with 10 seeds and all runs convey the same results.

\subsubsection{MNIST Domain}
In the following section, we describe the data, model, and optimization techniques for the experiments with the MNIST domain.

\begin{figure}[h]
    \centering
    \includegraphics[width=2in]{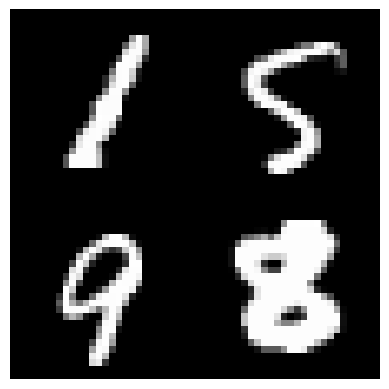}
    \caption{Example input from our MNIST regression task. This image represents the number $.1598$, which is then transformed by either $\sin$ or $\log$ to construct our subtask targets.}
    \label{fig:mnist-sample}
\end{figure}

\begin{figure*}[h]
    \begin{subfigure}{0.15\textwidth}
        \centering
            \includegraphics[width=\textwidth]{figures_v2/color_legend.png}
    \end{subfigure}
    \hfill
    \begin{subfigure}{0.4\textwidth}
        \centering
            \includegraphics[width=\textwidth]{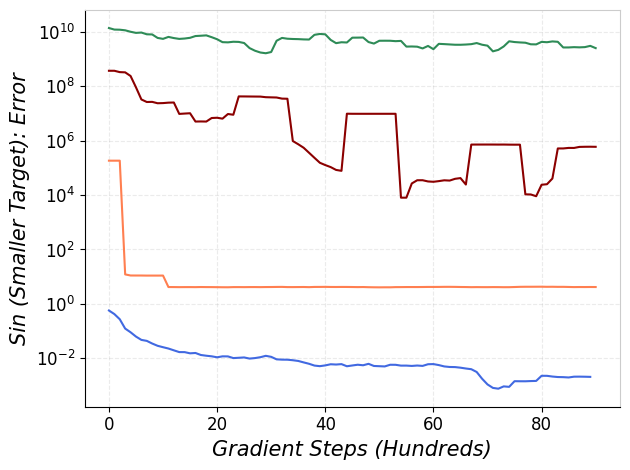}
        \caption{Error of samples with $\sin$ target: smaller target scale.}
        \label{fig:mnist-small-log-l}
    \end{subfigure}
    \hfill
    \begin{subfigure}{0.4\textwidth}
        \centering
        \includegraphics[width=\textwidth]{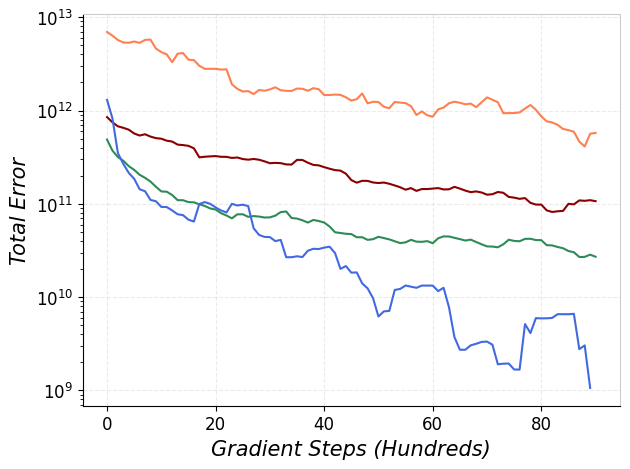}
        \caption{Total error of all samples including targets from both $\log$ and $\sin$ functions.}
        \label{fig:mnist-log-s-sin-l-b}
    \end{subfigure}
    \hfill
    \caption{Figure \ref{fig:mnist-log-s-sin-l-b} is the total test-set error of the MNIST dataset regression task with a single target derived from two subtasks: $\sin$ function with a mean value of $1$ and $\log$ function with a mean value of $9 \times 10^6$. The MNIST image input pixels are inverted if the target is the $\sin$ function and uninverted if the target is the $\log$ function. Figure \ref{fig:mnist-small-log-l} is the test-set error of the subset of samples having a target that is the $\sin$ function (smaller scale) during the same training process.}
    \label{fig:mnist-sin-s-log-l}
\end{figure*}

\begin{figure*}[h]
    \begin{subfigure}{0.15\textwidth}
    \centering
        \includegraphics[width=\textwidth]{figures_v2/color_legend.png}
    \end{subfigure}
    \hfill
    \begin{subfigure}{0.4\textwidth}
    \centering
        \includegraphics[width=\textwidth]{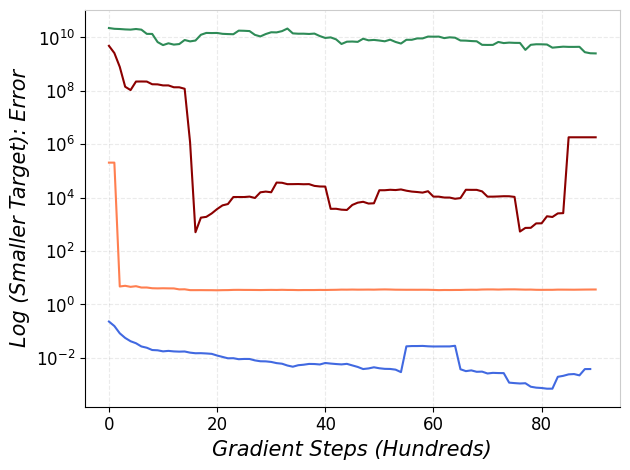}
        \caption{Error of samples with $\log$ target: smaller target scale.}
        \label{fig:mnist-small-log-s}
    \end{subfigure}
    \hfill
    \begin{subfigure}{0.4\textwidth}
        \centering
        \includegraphics[width=\textwidth]{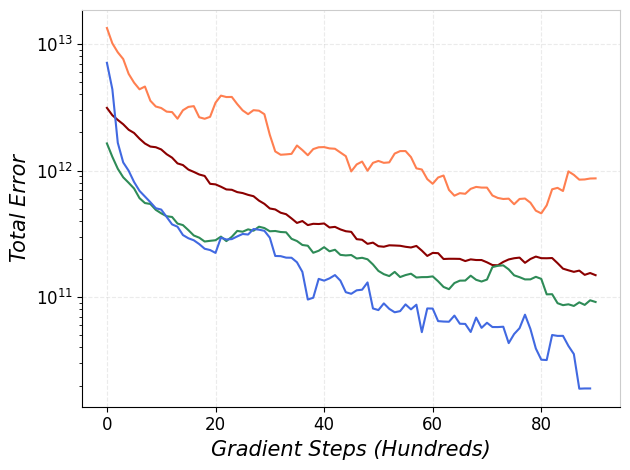}
        \caption{Total error of all samples including targets from both $\log$ and $\sin$ functions.}
        \label{fig:mnist-sin-l-log-s-b}
    \end{subfigure}
    \hfill
    \caption{Figure \ref{fig:mnist-sin-l-log-s-b} is the total test-set error of the MNIST dataset regression task with a single target derived from two subtasks: $\log$ function with a mean value of $1.67$ and $\sin$ function with a mean value of $18 \times 10^6$. The MNIST image input pixels are inverted if the target is the $\sin$ function and uninverted if the target is the $\log$ function. Figure \ref{fig:mnist-small-log-s} is the test-set error of the subset of samples having a target that is the $\log$ function (smaller scale) during the same training process.}
    \label{fig:mnist-sin-l-log-s}
\end{figure*}

\begin{figure*}[h]
    \begin{subfigure}{0.15\textwidth}
    \centering
        \includegraphics[width=\textwidth]{figures_v2/color_legend.png}
    \end{subfigure}
    \hfill
    \begin{subfigure}{0.4\textwidth}
    \centering
        \includegraphics[width=\textwidth]{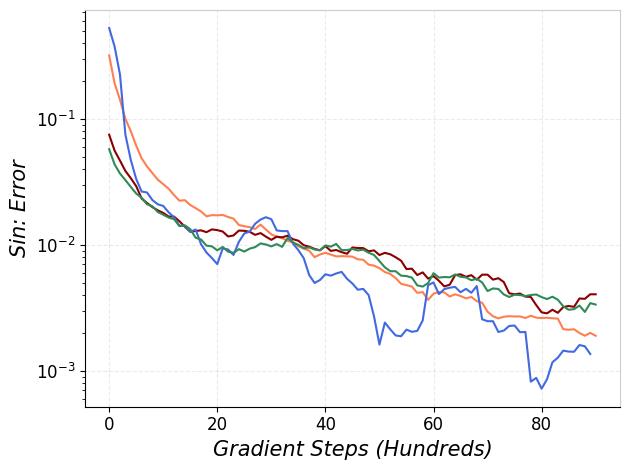}
        \caption{Error of $\sin$ subtask.}
        \label{fig:mnist-both-sin-s}
    \end{subfigure}
    \hfill
    \begin{subfigure}{0.4\textwidth}
        \centering
        \includegraphics[width=\textwidth]{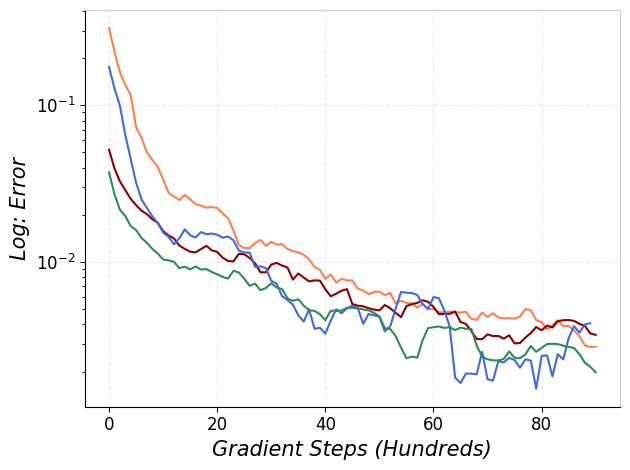}
        \caption{Error of $\log$ subtask.}
        \label{fig:mnist-both-log-s}
    \end{subfigure}
    \hfill
    \caption{The figure above depicts the test-set error during training of the 1-dimensional toy dataset regression task with a single target derived from two subtasks: $\log$ function with a mean value of $1.67$ and $\sin$ function with a mean value of $1$. The MNIST image input pixels are inverted if the target is the $\sin$ function and uninverted if the target is the $\log$ function. Figure \ref{fig:mnist-both-sin-s} is the test-set error of the subset of samples having a target that is the $\sin$ function. Figure \ref{fig:mnist-both-log-s} is the test-set error of the subset of samples having a target that is the $\log$ function during the same training process.}
    \label{fig:mnist-sin-s-log-s}
\end{figure*}

\subparagraph{Dataset}
The input to the model is the concatenation of $4$ randomly sampled images from the MNIST dataset. Each image in the MNIST dataset is $28 \times 28$ pixels. Because we concatenate $4$ randomly sampled images into a square, the input to the model is a $56\times 56$ pixel image - see Figure \ref{fig:mnist-sample}. Each pixel was normalized by the mean and standard deviation of the dataset. The 4 randomly sampled digits are concatenated and then divided by $10,000$ to produce a value between $(0-1)$. This value is then applied to another function to produce a target for a subtask. All models were trained with $10,000,000$ samples with a batch size of $1,000$, resulting in $10,000$ gradient steps. The experiment in Figure \ref{fig:mnist-sin-s-log-l} corresponds to a model trained on subtasks $\sin_s$ and $\log_l$, the experiment in Figure \ref{fig:mnist-sin-l-log-s} corresponds to a model trained on $\log_s$ and $\sin_l$, and the experiment in Figure \ref{fig:mnist-sin-s-log-s} corresponds to a model trained on $\sin_s$ and $\log_s$. Samples with targets having a $\sin$ function had the pixels in the input inverted (from black to white or white to black). Samples with targets having a $\log$ function were kept in its original form. In a single training batch, each sample has an equal probability of having a target be a $\sin$ or $\log$ function.

\subparagraph{Model and Optimization}
Each Figure (\ref{fig:mnist-sin-s-log-l}, \ref{fig:mnist-sin-l-log-s}, \ref{fig:mnist-sin-s-log-s}) presents the error for 2 subtasks over the course of training for a model trained with MSE, a model trained with MSE including target standardization, a model trained with MAE, and a model trained with autoregressive regression. All models were optimized via the Adam optimizer \citet{https://doi.org/10.48550/arxiv.1412.6980}. Learning rates for presented results were chosen using the same methodology for all types of models: all half orders of magnitude from $10^1 \rightarrow 10^{-6}$ were tested and the best performing learning rate was chosen. The autoregressive models were able to be trained with the same learning rate regardless of target scale, with a learning rate of $.0005$, whereas the MSE and MAE models required varying learning rates dependent on the scale of the targets. The model architecture is a convolutional neural network \citet{lecun1995convolutional} with $2$ convolutional layers including a batch normalization and rectified linear unit activation function followed by $5$ linear layers of dimension $1,024$ with rectified linear unit activation function. The model trained with autoregressive regression has $8$ autoregressive steps each with a bin size of $100$. Training was executed on a RTX A6000 GPU. Each experiment was run with 10 seeds and all runs convey the same results.

\subsubsection{Amazon Review Dataset}
In the following section, we describe the data, model, and optimization techniques for the experiments with the Amazon review dataset.
\subparagraph{Dataset}
The Amazon review datset includes $233.1$ million reviews of products including rating score, text, and product metadata. Using the text review, we created a byte pair encoding \citet{bpe} with vocab size of $10,000$. We define rating as the product review score divided by the max rating score of $5$, resulting in a value in $[0, 1]$. We define the number of characters as the number of characters in the review text. To fit longer reviews into memory and to make each task more challenging, we included the first $500$ tokens of the review as input to the model. The dataset is shuffled and a held out portion of the data is used to compute the error reported in Figure \ref{fig:amazon-review}. A batch size of $512$ is used with a total of $51,200,000$ samples resulting in $100,000$ gradient update steps. Samples with a Rating target had an additional unique token appended to the review, while samples with a Number of Characters target had a different unique token appended to the review. In a single training batch, each sample has an equal probability of having a target be a the review Rating or the review Number of Characters.

\subparagraph{Model and Optimization}
Both models were optimized via the Adam optimizer \citet{https://doi.org/10.48550/arxiv.1412.6980}. Learning rates for presented results were chosen using the same methodology for all types of models: all half orders of magnitude from $10^1 \rightarrow 10^{-6}$ were tested and the best performing learning rate was chosen. The model trained with autoregressive regression has $8$ autoregressive steps with a bin size of $100$. Model architecture is a transformer network \citep{vaswani2017attention} with $4$ heads, $4$ encoder layers, $4$ decoder layers, with a dimension of $512$. Training was executed on a RTX A6000 GPU. Each experiment was run with 10 seeds and all runs convey the same results.

\subsubsection{Learning Rate Stability}
In the following section, we describe the data, model, and optimization techniques for the Learning Rate Stability Experiments.
\subparagraph{Dataset}
The MNIST data set described in the MNIST Domain section was used. To generate different orders of magnitude, we used the following function to generate the targets:
\begin{equation}
    \sin_{m}(x) = 3\cdot 10^m\cdot (\sin(2\pi \cdot x) + 1) + 5\cdot 10^m
\end{equation}
Where $m$ denotes the order of magnitude of the target and $x$ is the normalized concatenated integers in the MNIST samples. All experiments were trained with $51,200,000$ samples with a batch size of $512$, resulting in $100,000$ gradient steps.

\subparagraph{Model and Optimization}
Both models were optimized via the Adam optimizer \citet{https://doi.org/10.48550/arxiv.1412.6980}. The model architecture is a convolutional neural network \citet{lecun1995convolutional} with $2$ convolutional layers including a batch normalization and rectified linear unit activation function followed by $2$ linear layers of dimension $2,048$ with rectified linear unit activation function. The model trained with autoregressive regression has $10$ autoregressive steps each with a bin size of $10$.Various learning rates and order of magnitude of the targets affect the final error reported in Figure \ref{fig:lr-combined}. Training was executed on a RTX A6000 GPU. 

\subsubsection{Computational Tractability}
In the set of experiments reported in Figure \ref{fig:memcombined}, we report the memory usage on various autoregressive step sizes resulting in different number of buckets being expressed by the model. The dataset is the MNIST dataset described in the MNIST Domain section, with a batch size of 5. The model used is also the same as that described in the MNIST Domain section, except with varying number of autoregressive steps and number of bins in each step. The GPU memory usage was tracked via the NVIDIA system management interface.

\subsection{Error}
In the following section, we describe the error metric we plot on all figures. The error for all models presented in the figures is the mean squared error. Models trained via MSE and MAE output a point-estimate, whereas models trained via HL output a distribution. Ideally, to compare these two predictive methods, we would compute the mean of the distribution predicted by HL and compute its squared-error to the targets. However, given that autoregressive regression outputs a distribution with an exponential number of bins in the size of its output, it is generally infeasible to compute the mean in closed form. Instead, we approximate the mean by taking $100$ samples from the output distribution. Note that this is an upper bound to the error as measured against the true mean.

\subsection{Additional Experiment Figures}

\end{document}